# CONCURRENT LEXICALIZED DEPENDENCY PARSING: THE *ParseTalk* MODEL

## Norbert Bröker, Udo Hahn & Susanne Schacht

$\mathcal{CLIF}$- Computational Linguistics Research Group
Freiburg University
D-79085 Freiburg, Germany

email: {nobi, hahn, sue}@coling.uni-freiburg.de

*Abstract.* A grammar model for concurrent, object-oriented natural language parsing is introduced. Complete lexical distribution of grammatical knowledge is achieved building upon the head-oriented notions of valency and dependency, while inheritance mechanisms are used to capture lexical generalizations. The underlying concurrent computation model relies upon the actor paradigm. We consider message passing protocols for establishing dependency relations and ambiguity handling.

## 1 INTRODUCTION

In this paper, we propose a grammar model that combines lexical organization of grammatical knowledge with lexicalized control of the corresponding parser in an object-oriented specification framework. Recent developments in the field of linguistic grammar theory have already yielded a rigid *lexical modularization*. This fine-grained decomposition of linguistic knowledge can be taken as a starting point for lexicalized control. Current lexicalized grammars (for instance, HPSG: Pollard & Sag, 1987; CG: Hepple, 1992; Lexicalized TAG: Schabes, Abeille & Joshi, 1988), however, still consider lexical items as passive data containers whose content is uniformly interpreted by global control mechanisms (e.g., unification, functional composition, tree adjunction). Diverging from these premises, we assign full procedural autonomy to lexical units and treat them as active *lexical processes* communicating with each other by message passing. Thus, they dynamically establish heterogeneous communication lines in order to determine each lexical item's functional role. While the issue of lexicalized control has early been investigated in the paradigm of conceptual parsing (Riesbeck & Schank, 1978), and word expert parsing in particular (Small & Rieger, 1982), these proposals are limited in several ways. First, they do not provide any general mechanism for the systematic incorporation of grammatical knowledge. Second, they do not supply any organizing facility to formulate generalizations over sets of lexical items. Third, lexical communication is based on an entirely informal protocol that lacks any grounding in principles of distributed computing.

We intend to remedy these methodological shortcomings by designing a radically lexicalized grammar on the basis of *valency* and *dependency* (these head-oriented notions already figure in different shapes in many modern linguistic theories, e.g., as subcategorizations, case frames, theta roles), by introducing *inheritance* as a major organizational mechanism (for a survey of applying inheritance in modern grammar theory, cf. Daelemans, De Smedt & Gazdar, 1992), and by specifying a *message passing* protocol that is grounded in the actor computation model (Agha & Hewitt, 1987). As this protocol allows for asynchronous message passing, *concurrency* enters as a theoretical notion at the level of grammar specification, not only as an implementational feature. The *ParseTalk* model outlined in this paper can therefore be considered as an attempt to replace the static, global-control paradigm of natural language processing by a dynamic, local-control model.

The design of such a grammar and its associated parser responds to the demands of complex *language performance* problems. By this, we mean understanding tasks, such as large-scale text or speech understanding, which not only require considerable portions of grammatical knowledge but also a vast amount of so-called non-linguistic, e.g., domain and discourse knowledge. A major problem then relates to the interaction of the different knowledge sources involved, an issue that is not so pressing when monolithic grammar knowledge essentially boils down to syntactic regularities. Instead of subscribing to any serial model of control, we build upon evidences from computational text understanding studies (Granger, Eiselt & Holbrook, 1986; Yu & Simmons, 1990) as well as psycholinguistic experiments, in particular those worked out for the class of interactive language processing models (Marslen-Wilson & Tyler, 1980; Thibadeau, Just & Carpenter, 1982). They reveal that various knowledge sources are accessed in an *a priori* unpredictable order and that a significant amount of *parallel* processing occurs at various stages of the (human) language processor. Therefore, computationally and cognitively plausible models of natural language understanding should account for parallelism at the *theoretical* level of language description. Currently, *ParseTalk* provides a specification platform for computational *language performance modeling*.[1] In the future, this vehicle can be used as a testbed for the configuration of cognitively adequate parsers. Moving performance considerations to the level of grammar design is thus in strong contrast to any competence-based account which assigns structural well-formedness conditions to the grammar level and leaves their computation to (general-purpose) parsing algorithms, often at the cost of vast amounts of ambiguous structural descriptions.

---

[1] We only mention that performance issues become even more pressing when natural language understanding tasks are placed in real-world environments and thus additional complexity is added by ungrammatical natural language input, noisy data, as well as lexical, grammatical, and conceptual specification gaps. In these cases, not only multiple knowledge sources have to be balanced but additional processing strategies must be supplied to cope with these phenomena in a robust way. This places extra requirements on the integration of *procedural* linguistic knowledge within a performance-oriented language analysis framework, viz. strategic knowledge how to handle incomplete or faulty language data and grammar specifications.

## 2 *ParseTalk*'s CONCEPTUAL FRAMEWORK

The *ParseTalk* model is based on a fully lexicalized grammar. Grammatical specifications are given in the format of *valency constraints* attached to each lexical unit, on which the computation of concrete *dependency relations* is based. By way of inheritance the entire collection of lexical items is organized in *lexical hierarchies* (these constitute the *lexical grammar*), the lexical items forming their leaves and the intermediary nodes representing grammatical generalizations in terms of word classes. This specification is similar to various proposals currently investigated within the unification grammar community (Evans & Gazdar, 1990). The concurrent computation model builds upon and extends the formal foundations of the *actor model*, a theory of object-oriented computation that is based on asynchronous message passing.

### 2.1 The Grammar Model

The grammar model underlying the *ParseTalk* approach considers dependency relations between words as the fundamental notion of linguistic analysis. A *modifier* is said to depend on its *head* if the modifier's occurrence is permitted by the head but not *vice versa*[2]. *Dependencies* are thus asymmetric binary relations that can be established by local computations involving only two lexical items; they are tagged by dependency relation names from the set $\mathcal{D} = \{$spec, subj, ppatt, ...$\}$[3]. Co-occurrence restrictions between lexical items are specified as sets of *valencies* that express various constraints a head places on permitted modifiers. These constraints incorporate the following descriptive units:

1. **categorial:** $\mathcal{C} = \{$WordActor, Noun, Substantive, Preposition, ...$\}$ denotes the set of word classes, and $isa_\mathcal{C} = \{$(Noun, WordActor), (Substantive, Noun), (Preposition, WordActor), ...$\} \subset \mathcal{C} \times \mathcal{C}$ denotes the subclass relation yielding a hierarchical ordering in $\mathcal{C}$ (cf. also Fig.1).

2. **morphosyntactic:** A unification formalism (similar in spirit to Shieber, 1986) is used to represent morphosyntactic regularities. It includes atomic terms from the set $\mathcal{T} = \{$nom, acc, ..., sg, pl, ...$\}$, complex terms associating labels from the set $\mathcal{L} = \{$case, num, agr, ...$\} \cup \mathcal{D}$ with embedded terms, value disjunction (in curly braces), and coreferences (numbers in angle brackets). $\mathcal{U}$ denotes the set of allowed feature structures, $\nabla$ the unification operation, $\bot$ the inconsistent element. Given $u \in \mathcal{U}$ and $l \in \mathcal{L}$, the *expansion* [l : u] denotes the complex term containing only one label, l, with value u. If u is a complex term containing l at top level, the *extraction* u\l is defined to be the value of l in u. By definition, u\l yields $\bot$ in all other cases.

3. **conceptual:** The concept hierarchy consists of a set of concept names $\mathcal{F} = \{$Hardware, Computer, Notebook, Harddisk, ...$\}$ and a subclass relation $isa_\mathcal{F} = \{$(Computer, Hardware), (Notebook, Computer), ...$\} \subset \mathcal{F} \times \mathcal{F}$. The set of conceptual role names $\mathcal{R} = \{$HasPart, HasPrice, ...$\}$ contains labels of possible conceptual relations (a frame-style, classification-based knowledge representation model in the spirit of MacGregor (1991) is assumed). The relation $cic \subseteq \mathcal{F} \times \mathcal{R} \times \mathcal{F}$ implements conceptual integrity constraints: $(f, r, g) \in cic$ iff any concept subsumed by $f \in \mathcal{F}$ may be modified by any concept subsumed by $g \in \mathcal{F}$ in relation $r \in \mathcal{R}$, e.g, (Computer, hasPart, Harddisk) $\in cic$. From $cic$ the relation $permit = \{(x, r, y) \in \mathcal{F} \times \mathcal{R} \times \mathcal{F} \mid \exists f, g \in \mathcal{F} : (f, r, g) \in cic \wedge x\, isa_\mathcal{F}^*\, f \wedge y\, isa_\mathcal{F}^*\, g\}$ (* denotes the transitive closure) can be derived which explicitly states the range of concepts that can actually be related. For brevity, we restrict this exposition to the attribution of concepts and do not consider quantification, etc. (cf. Creary & Pollard, 1985).

4. **ordering:** The (word-class specific) set $order \subset \mathcal{D}^n$ contains n-tuples which express ordering constraints on the valencies of each word class. Legal orders of modifiers must correspond to an element of $order$. The (word specific) function $occurs : \mathcal{D} \to \mathcal{N}_0$ associates dependency names with the modifier's (and self's) text position (0 for valencies not yet occupied). Both specifications appear at the lexical head only, since they refer to the head and all of its modifiers.

With these definitions, a valency can be characterized as an element of the set $\mathcal{V} \subset \mathcal{D} \times \mathcal{C} \times \mathcal{U} \times \mathcal{R}$. Focusing on one dependency relation from the example *"Compaq entwikkelt einen Notebook mit einer 120-MByte-Harddisk"* [*"Compaq develops a notebook with a 120-MByte hard disk"*], the above criteria are illustrated in Table 1. The *feature* structure of the two heads, *"mit"* and *"Notebook"*, is given prior to and after the establishment of the dependency relation. The *concepts* of each of the phrases, 120MB-HARDDISK-00004 and NOTEBOOK-00003, are stated. The *order* constraint of *"Notebook"* says that it may be preceded by a specifier (spec) and attributive adjectives (attr), and that it may be followed by prepositional phrases (ppatt). The valency for prepositional phrases described in the last row states which *class*, *feature*, and *domain* constraints must be fulfilled by candidate modifiers.

The predicate SATISFIES (cf. Table 2) holds when a candidate modifier fulfills the constraints stated in a specified valency of a candidate head. If SATISFIES evaluates to true, a dependency valency.name is established (object.attribute denotes the value of the property attribute at object). As can easily be verified, SATISFIES is fulfilled for the combination of *"mit"*, the prepositional valency, and *"Notebook"* from Table 1.

---
[2] Although phrases are not explicitly represented (e.g., by non-lexical categories), we consider each complete subtree of the dependency tree a phrase (this convention allows discontinuous phrases as well). A dependency is not treated as a relation between words (as in Word Grammar (Hudson, 1990, p.117), but between a word and a dependent phrase (as in Dependency Unification Grammar (Hellwig, 1988)). The root of a phrase is taken to be the representative of the whole phrase.

[3] Additionally, $\mathcal{D}$ contains the symbol self which denotes the currently considered lexical item. This symbol occurs in feature structures (see 2. below) and in the ordering relations *order* and *occurs* (4. below).

| | Attributes | Lexical items (head underlined) | prior to dependency establishment | after dependency establishment |
|---|---|---|---|---|
| Possible Modifier | class ∈ $\mathcal{C}$<br><br>features ∈ $\mathcal{U}$<br><br><br>concept ∈ $\mathcal{F}$<br>position ∈ $\mathcal{N}$ | <u>mit</u> einer 120-MByte-Harddisk | Preposition<br>$\begin{bmatrix} \text{self} & [\text{form mit}] \\ \text{pobj} & \text{agr} \begin{bmatrix} \text{case dat} \\ \text{gen fem} \\ \text{num sg} \end{bmatrix} \end{bmatrix}$<br>120MB-HARDDISK-00004<br>5 | Preposition<br>$\begin{bmatrix} \text{self} & [\text{form mit}] \\ \text{pobj} & \text{agr} \begin{bmatrix} \text{case dat} \\ \text{gen fem} \\ \text{num sg} \end{bmatrix} \end{bmatrix}$<br>120MB-HARDDISK-00004<br>5 |
| Possible Head | features ∈ $\mathcal{U}$<br><br><br><br>concept ∈ $\mathcal{F}$<br>$order \subset \mathcal{D}^n$<br>$occurs : \mathcal{D} \to \mathcal{N}_0$ | einen <u>Notebook</u> | $\begin{bmatrix} \text{self} & \text{agr} <1> = \begin{bmatrix} \text{case acc} \\ \text{gen mas} \\ \text{num sg} \end{bmatrix} \\ \text{spec} & [\text{agr} <1>] \end{bmatrix}$<br>NOTEBOOK-00003<br>{<spec, attr, self, ppatt>}<br>{(spec, 3), (attr, 0), (self, 4), (ppatt, 0)} | $\begin{bmatrix} \text{self} & \text{agr} <1> = \begin{bmatrix} \text{case acc} \\ \text{gen mas} \\ \text{num sg} \end{bmatrix} \\ \text{spec} & [\text{agr} <1>] \\ \text{ppatt} & [\text{form mit}] \end{bmatrix}$<br>NOTEBOOK-00003<br>{<spec, attr, self, ppatt>}<br>{(spec, 3), (attr, 0), (self, 4), (ppatt, 5)} |
| Valencies (only one of the set is considered) | name ∈ $\mathcal{D}$<br>class ∈ $\mathcal{C}$<br>features ∈ $\mathcal{U}$<br>domain ⊆ $\mathcal{R}$ | | ppatt<br>Preposition<br>$[\text{ppatt} \; [\text{form mit}]]$<br>{HasHarddisk, HasPrice, ...} | not applicable |

**TABLE 1. An illustration of grammatical specifications in the *ParseTalk* model**

```
SATISFIES (modifier, valency, head) :⇔
    modifier.class isa*_C valency.class
  ∧ (([valency.name:(modifier.features\self)] ∇ valency.features)
        ∇ head.features) ≠ ⊥
  ∧ ∃ role ∈ valency.domain :
        (head.concept, role, modifier.concept) ∈ permit
  ∧ ∃ <d_1,... d_n> ∈ head.order : ∃ k ∈ {1, ..n} :
        (valency.name = d_k
        ∧ (∀ 1 ≤ i < k : (head.occurs (d_i) < modifier.position))
        ∧ (∀ k < i ≤ n : (head.occurs (d_i) = 0
                            ∨ head.occurs (d_i) > modifier.position))
```
**TABLE 2. The SATISFIES predicate**

Note that unlike most previous dependency grammar formalisms (Starosta & Nomura, 1986; Hellwig, 1988; Jäppinen, Lassila & Lehtola, 1988; Fraser & Hudson, 1992) this criterion assigns equal opportunities to syntactic as well as conceptual conditions for computing valid dependency relations. Information on word classes, morphosyntactic features, and order constraints is purely syntactic, while conceptual compatibility introduces an additional description layer to be satisfied before a grammatical relation may be established (cf. Muraki, Ichiyama & Fukumochi, 1985; Lesmo & Lombardo, 1992). Note that we restrict the scope of the unification module in our framework, as only morphosyntactic features are described using this subformalism. This contrasts sharply with standard unification grammars (and with designs for dependency parsing as advocated by Hellwig (1988) and Lombardo (1992)), where virtually all information is encoded in terms of the unification formalism[4].

### 2.1.1 A Look at Grammatical Hierarchies

The grammatical specification of a lexical entry consists of structural criteria (valencies) <u>and</u> behavioral descriptions (protocols). In order to capture relevant generalizations and to support easy maintenance of grammar specifications, both are represented in hierarchies (cf. Genthial, Courtin & Kowarski (1990) and Fraser & Hudson (1992) for inheritance that is restricted to structural criteria). The *valency hierarchy* assigns valencies to lexemes. We will not consider it in depth here, since it captures only traditional grammatical notions, like transitivity or reflexivity. The organizing principle is the subset relation on valency sets. The *word class hierarchy* contains word class specifications that cover distributional and behavioral properties. Fig. 1 illustrates the behavioral criterion by defining for each class different messages (the messages for WordActor are discussed in Sections 3 and 4). Within the Noun part of the word class hierarchy, there are different methods for anaphora resolution reflecting different structural constraints on possible antecedents for nominal anaphora, reflexives and personal pronouns. The word class hierarchy cannot be generated automatically, since classification of program specifications (communication protocols, in our case) falls out of the scope of state-of-the-art classifier algorithms. On the other hand, the *concept hierarchy* is based on the subsumption relation holding between concepts, which is computed by a terminological classifier. Most lexicon entries refer to a corresponding domain concept and thus allow conceptual restrictions to be checked.

---
[4] Typed unification formalisms (Emele & Zajac, 1990) would easily allow for the integration of word class information. Ordering constraints and conceptual restrictions (such as value range restrictions or elaborated integrity constraints), however, are not so easily transferable, because, e.g., the conceptual constraints go far beyond the level of atomic semantic features still prevailing in unification formalisms.

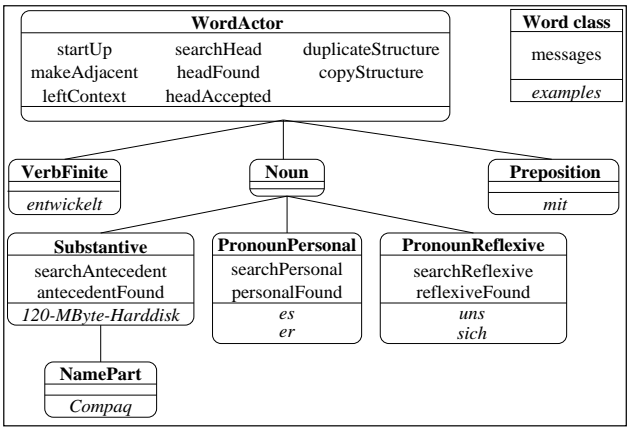

**FIGURE 1. Fragment of the word class hierarchy**

### 2.2 The Actor Computation Model

The actor model of computation combines object-oriented features with concurrency and distribution in a methodologically clean way. It assumes a collection of independent objects, the *actors*, communicating via asynchronous message passing. An actor can send messages only to other actors it knows about, its *acquaintances*. The arrival of a message at an actor is called an *event*; it triggers the execution of a method that is composed of atomic actions, viz. creation of new actors (<u>create</u> actorType (acquaintances)), sending of messages to acquainted or a newly created actors (<u>send</u> actor message), or specification of new acquaintances (<u>become</u> (acquaintances)). An actor system is dynamic, since new actors can be created and the communication topology is reconfigurable. We assume actors that process a single message at a time, step by step (Hewitt & Atkinson, 1979). For convenience, we establish a synchronous request-reply protocol (Lieberman, 1987) to compute functions such as unification of feature structures and queries to a (conceptual) knowledge base. In contrast to simple messages which unconditionally trigger the execution of a method at the receiving actor, we define *complex word actor messages* as full-fledged actors with independent computational abilities. Departure and arrival of complex messages are actions which are performed by the message itself, taking the sender and the target actors as parameters. Upon arrival, a complex message determines whether a copy is forwarded to selected acquaintances of its receiver and whether the receiver may process the message on its own (cf. Schacht, Hahn & Bröker (1994) for a treatment of the parser's behavioral aspects).

The following syntax elements will be used subsequently: a *program* contains *actor definitions* (declaring the acquaintances and defining the methods of actors instantiated from this definition) and *actor message definitions* (stating distribution and computation *conditions*). Method definitions contain the *message key*, the formal *parameters* and a composite *action*:

```
actorDef   ::= defActor actorType (acquaintance*)
                   methodDef*
methodDef  ::= meth messageKey (param*) (action)
messDef    ::= defMsg messageType (acquaintance*)
                   (((if condition distributeTo tag))*
                   if condition compute
                   ((if condition distributeTo tag))*)
action     ::= action; action
             | if condition (action) [ else (action) ]
             | send actor messageKey (param*)
             | become (acquaintance*)
             | create actorType (acquaintance*)
             | for var in set : (action)
```

condition is a locally computable predicate, written as PREDICATE (actor*); actor stands for acquaintances, parameters, newly created actors, the performing actor itself (<u>self</u>) or the undefined value (<u>nil</u>); actor.acquaintance yields the corresponding acquaintance of actor; <u>for</u> var <u>in</u> set: (action) evaluates action for each element of set.

## 3 A SIMPLIFIED PROTOCOL FOR ESTABLISHING DEPENDENCY RELATIONS

The protocol described below allows to establish dependency relations. It integrates structural restrictions on dependency trees and provides for domesticated concurrency

### 3.1 Synchronizing Actor Activities: Reception Protocol

A reception protocol allows an actor to determine when all events (transitively) caused by a message have terminated. This is done by sending replies back to the initiator of the message. Since complex messages can be quasi-recursively forwarded, the number of replies cannot be determined in advance. In addition, each actor receiving such a message may need an arbitrary amount of processing time to terminate the actions caused by the message (e.g., the establishment of a dependency relation requires communication via messages that takes indeterminate time). Therefore, each actor receiving the message must reply to the initiator once it has terminated processing, informing the initiator to which actors the message has been forwarded.

A message is a *reception message* if (1) the receiver is required to (asynchronously) reply to the initiator with a receipt message, and (2) the initiator queues a reception task. An *(explicit) receipt message* is a direct message containing a set of actor identities as a parameter. This set indicates to which actors the reception message has been forwarded or delegated. The enclosed set enables the receiver (which is the initiator of the reception message) to wait until all receipt messages have arrived[5]. In addition to explicit receipts, which are messages solely used for termination detection, there are regular messages that serve a similar purpose besides their primary function within the parsing process. They are called *implicit receipt messages* (one example is the **headAccepted** message described in Section 3.3). A *reception task* consists of a set of partial descriptions of the messages that must be received (implicit as well as explicit), and an action to be executed after all receipts have arrived (usually, sending a message).

---

[5] This, of course, only happens if the distribution is limited: The **search-Head** message discussed below is only distributed to the head of each receiver, which must occur in the same sentence. This ensures a finite actor collection to distribute the message to, and guarantees that the reception task is actually triggered.

```
defActor wordActor (head deps vals feats ...)        # head, dependencies, valencies, and features acquaintances
  meth searchHead (sender target init)               # processed at candidate heads (compute from the message definition)
    (for val in vals:                                # check all valencies of the possible head
      (if SATISFIES (init val self)                  # valency check adapted from Table 2
        (send (create headFound (self init val.name feats\val.name)) depart;  # reply to initiator, imposing restrictions
         become (head deps vals (feats ∇ init.feats) ...)  # expand grammatical description of head
        else (send (create receipt (self init {head})) depart)))  # send a receipt with the head the message was forwarded to
                                                     # depart realizes the departure of a complex message
  meth headFound (sender target name headFeats)      # processed at the initiator of a searchHead message
    (send (create headAccepted (self sender name)) depart);  # reply to head
     become (sender deps vals (feats ∇ headFeats) ...))  # store sender as head of self, restrict self's features
  meth headAccepted (modifier target name)           # processed at the head only
    (for dep in deps:                                # check all dependencies
      (if (name = dep.name)                          # relation name is identical
        (send dep store (modifier))))               # send the dependency the message store to store the modifier
     send (create receipt (self modifier {head})) depart)  # send a receipt with the head the message was forwarded to
```

**TABLE 3. Method definitions for searchHead, headFound, headAccepted**

## 3.2 Encoding Structural Restrictions

Word actors conduct a bottom-up search for possible heads; the principle of non-crossing arcs (*projectivity* of the dependency tree) is guaranteed by the following forwarding mechanism. Consider the case of a newly instantiated word actor $w_n$ searching its head to the left (the opposite direction is handled in a similar way). In order to guarantee projectivity one has to ensure that only word actors occupying the outer fringe of the dependency structure (between the current absolute head $w_j$ and the rightmost element $w_{n-1}$) receive the search message of $w_n$ (these are circled in Fig. 2)[6]. This forwarding scheme is reflected in the following simplified message definition:

```
defMsg searchHead (sender target initiator)
  ((if GOVERNED (target) distributeTo head)
        # forward a copy to head, identified by head ∈ D
    if true compute)
        # the message is always processed at the target;
        # the computation event is concretized in the word
        # actor specification in Table 3
```

Thus, a message searching for a head of its initiator is locally processed at each actor receiving it, and is forwarded to the head of each receiver, if one already exists.

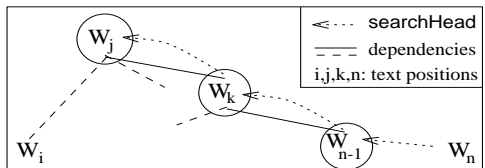

**FIGURE 2. Forwarding a search message**

Additionally, *direct messages* are used to establish a dependency relation. They involve no forwarding and may be specified as follows:

```
defMsg <directMessage> (sender target ...)
  (if true compute)
        # a direct message is always processed at the
        # target, no distribution condition can apply
```

Below, a number of messages of this type are used for negotiating dependencies, e.g., **headFound**, **headAccepted**, **receipt** (each with different parameters, as represented by "..." above).

## 3.3 An Excerpt from the Word Actor Script

The protocol for bottom-up establishment of dependencies consists of three steps: The search for a head (**searchHead**), the reply of a suitable head to the initiator of the search (**headFound**), and the acceptance by the initiator (**headAccepted**), thereby becoming a modifier of the head. The corresponding method definitions are given in Table 3 (note that these methods are defined for one actor type here, but are executed by different actors during parsing). The protocol allows alternative attachments to be checked *concurrently*, since each actor receiving **searchHead** may process it locally, while the message is simultaneously distributed to its head.

The specification of methods as above gives a local view of an actor system, stating how each actor behaves when it receives a message. For a global view taking the actors' interaction patterns into account, cf. Schacht, Hahn & Bröker (1994).

## 4 AMBIGUITY HANDLING

There are two alternative processing strategies for ambiguities, viz. serial vs. parallel processing. We here focus on a parallel mode, specifying only necessary serializations. Whenever an ambiguity is detected, additional actors are created to represent different readings. The standard three-step negotiation scheme for dependencies can easily be accommodated to this duplication process. When a word actor receives the second (or n-th) **headFound** message it does not immediately reply with a **headAccepted** message, but initiates the copying of itself, its modifiers, and the prospective head (which, in turn, initiates copying its modifiers and head, if any). Copying modifiers proceeds by sending a **copyStructure** message to each actor involved, which evokes a (standard) **headAccepted** message returned by the actor copy. Copying the head is done via a **duplicateStructure** message, which will result in another **headFound** message to be returned. Since this **headFound** message is addressed to the ungoverned copy, the copy may reply as usual by sending a **headAccepted** message. Duplication of actors allows the concurrent processing of alternatives, and requires only limited overhead for the distribution of messages among duplicated actors.

---

[6] Additionally, $w_n$ may be governed by any word actor governing $w_j$, but due to the synchronization implemented by the receipt protocol, each head of $w_j$ must be located to the right of $w_n$.

### 4.1 Packing Ambiguities

Usually, a packed representation of ambiguous structures is preferred in the parsing literature (Tamura et al., 1991). This is feasible when syntactic analysis is the only determining factor for the distribution of partial structures. But if conceptual knowledge is taken into account, the distribution of a phrase is not fully determined by its syntactic structure. Possible conceptual relations equally influence the distribution of the phrase. Additionally, the inclusion of an ambiguous phrase in a larger syntactic context requires the modification of the conceptual counterparts. In a packed representation, there would have to be several conceptual counterparts, i.e., only the syntactic representation can be packed (and it might even be necessary to unpack it on-the-fly). Consequently, whenever conceptual analysis is integrated into the parsing process (as opposed to its interpretation in a later stage, thereby producing numerous ambiguities in the syntactic analysis), structure sharing is impossible, since different syntactic attachments result in different conceptual analyses, and no common structure is accessible that can be shared (cf. Akasaka (1991) for a similar argument). We expect that the overhead of duplication is compensated for by the ambiguity-reducing effects of integrating several knowledge sources.

### 4.2 Relation to Psycholinguistic Performance Models

It has been claimed that human language understanding proceeds in a more sequential mode, choosing one alternative and backtracking if that path fails (e.g., Hemforth, Konieczny & Strube, 1993). This model requires the ranking of all alternatives according to criteria referring to syntactic or conceptual knowledge. The protocol outlined so far could easily be accommodated to this processing strategy: All **headFound** messages must be collected, and the corresponding attachments ranked. The best attachment is selected, and only one **headAccepted** message sent. In case the analysis fails, the next-best attachment would be tried, until an analysis is found or no alternatives are left. Additionally, the dependencies established during a failed path would have to be released.[7]

## 5 COMPARISON TO RELATED WORK

The issue of object-oriented parsing and concurrency (for a survey, cf. Hahn & Adriaens, 1994) has long been considered from a purely *implementational* perspective. Message passing as an explicit control mechanism is inherent to various object-oriented implementations of standard rule-based parsers (cf. Yonezawa & Ohsawa (1988) for context-free and Phillips (1984) for augmented PSGs). Actor-based implementations are provided by Uehara et al. (1985) for LFGs and Abney & Cole (1986) for GB grammars. Similarly, a parallel implementation of a rule-based, syntax-oriented dependency parser has been described by Akasaka (1991). The consideration of concurrency at the grammar *specification* level has recently been investigated by Milward (1992) who properly relates notions from categorial and dependency grammar with a state logic approach, a formal alternative to the event-algebraic formalization underlying the *ParseTalk* model.

Almost any of these proposals lack serious accounts of the integration of syntactic knowledge with conceptual knowledge (cf. the end of Section 2.1 for similar considerations related to dependency grammars). The development of conceptual parsers (Riesbeck & Schank, 1978), however, was entirely dominated by conceptual expectations driving the parsing process and specifically provided no mechanisms to integrate linguistic knowledge into such a lexical parser in a systematic way. The pseudo-parallelism inherent to these early proposals, word expert parsing in particular (Small & Rieger, 1982), has in the meantime been replaced by true parallelism, either using parallel logic programming environments (Devos, Adriaens & Willems, 1988), actor specifications (Hahn, 1989) or a connectionist methodology (Riesbeck & Martin, 1986), while the lack of linguistic sophistication has remained.

A word of caution should be expressed regarding the superficial similarity between object-oriented and connectionist models. Connectionist methodology (cf. a survey by Selman (1989) of some now classical connectionist natural language parsing systems) is restricted in two ways compared with object-oriented computing. First, its communication patterns are determined by the hard-wired topology of connectionist networks, whereas in object-oriented systems the topology is flexible and reconfigurable. Second, the type and amount of data that can be exchanged in a connectionist network is restricted to *marker* and *value passing* together with severely limited computation logic (and-ing, or-ing of Boolean bit markers, determining maximum/minimum values, etc.), while none of these restrictions apply to *message passing* models. These considerations equally extend to *spreading activation* models of natural language parsing (Charniak, 1986; Hirst, 1987) which are not as constrained as connectionist models but less expressive than general message passing models underlying the object-oriented paradigm. As should be evident from the preceding exposition of the *ParseTalk* model, the complexity of the data exchanged and computations performed, in our case, require a full-fledged message-passing model.

## 6 CONCLUSIONS

The *ParseTalk* model of natural language understanding aims at the integration of a lexically distributed, dependency-based grammar specification with a solid formal foundation for concurrent, object-oriented parsing (cf. Hahn, Schacht & Bröker (forthcoming) for a more elaborated presentation). It conceives communication among and within different knowledge sources (grammar, domain and discourse knowledge) as the backbone for complex language understanding tasks. The main specification elements of the grammar model consist of categorial, mor-phosyntactic, conceptual, and ordering constraints in terms of valency specifications attached to single lexical items. The associated concurrent computation model is based on the actor paradigm of object-oriented programming. The *ParseTalk* model has been experimentally validated by a prototype system, a parser for German (for its implementational status, cf. Schacht, Hahn & Bröker, 1994).

---

[7] Note that all psycholinguistic studies we know of are referring to a constituency-based grammar model. Since our grammar is based on dependency relations, principles such as Minimal Attachment cannot be transferred without profound modification, since in a dependency tree the number of nodes is identical for all readings. Therefore, principles adapted to the structural properties of dependency trees must be formulated for preferential ranking.

**Acknowledgments**

The work reported in this paper is funded by grants from DFG (grants no. Ha 2097/1-1, Ha 2097/1-2) within a special research programme on cognitive linguistics. We like to thank our colleagues, P. Neuhaus, K. Schnattinger, M. Klenner, and Th. Hanneforth, for valuable comments and support.